 \documentclass[pmlr,twocolumn]{jmlr} 

\usepackage{booktabs}
\usepackage[load-configurations=version-1]{siunitx} 

\theorembodyfont{\upshape}
\theoremheaderfont{\scshape}
\theorempostheader{:}
\theoremsep{\newline}

\jmlrvolume{ML4H Extended Abstract Arxiv Index}
\jmlryear{2020}
\jmlrsubmitted{2020}
\jmlrpublished{}
\jmlrworkshop{Machine Learning for Health (ML4H) 2020} 

\newcommand{\mname}{\texttt{SnipBERT}\xspace}
\newcommand{\xhdr}[1]{\noindent{{\bf #1.}}}

\usepackage{lipsum} 
\theorembodyfont{\upshape} 

\usepackage{adjustbox}
\usepackage{multicol}
\usepackage{multirow}

\title[An Interpretable End-to-end Fine-tuning Approach for Long Clinical Text]{An Interpretable End-to-end Fine-tuning Approach \\ for Long Clinical Text}

\author{
\Name{Kexin Huang} \Email{kexinhuang@hsph.harvard.edu}\\
\addr Flatiron Health, Inc. New York, NY; Harvard University, Boston, MA\\
\Name{Sankeerth Garapati} \Email{sankeerth.garapati@flatiron.com}\\
\Name{Alexander S. Rich} \Email{arich@flatiron.com}\\
\addr Flatiron Health, Inc. New York, NY
}

\begin{document}

\maketitle
\vspace{-10mm}
\begin{abstract}
 Unstructured clinical text in EHRs contains crucial information for applications including decision support, trial matching, and retrospective research. Recent work has applied BERT-based models to clinical information extraction and text classification, given these models' state-of-the-art performance in other NLP domains. However, BERT is difficult to apply to clinical notes because it doesn't scale well to long sequences of text. In this work, we propose a novel fine-tuning approach called \mname. Instead of using entire notes, \mname identifies crucial snippets and then feeds them into a truncated BERT-based model in a hierarchical manner. Empirically, \mname not only has significant predictive performance gain across three tasks but also provides improved interpretability, as the model can identify key pieces of text that led to its prediction.
\end{abstract}

\section{Introduction}
\vspace{-2mm}

Electronic health records (EHRs) are an important source of data for applications like outcomes research, clinical trial matching, and decision support~\citep{shickel2017deep,khozin2017real}. A challenge in using EHR data for these purposes is that many patient characteristics, including diagnoses, treatments, and test results are often found only in unstructured text~\citep{weng2017medical,liu2018deep,boag2018s}. There is therefore increasing interest in using machine learning to automatically extract patient characteristics from EHR text.

\begin{figure*}
    \centering
    \includegraphics[width = 1.0\textwidth]{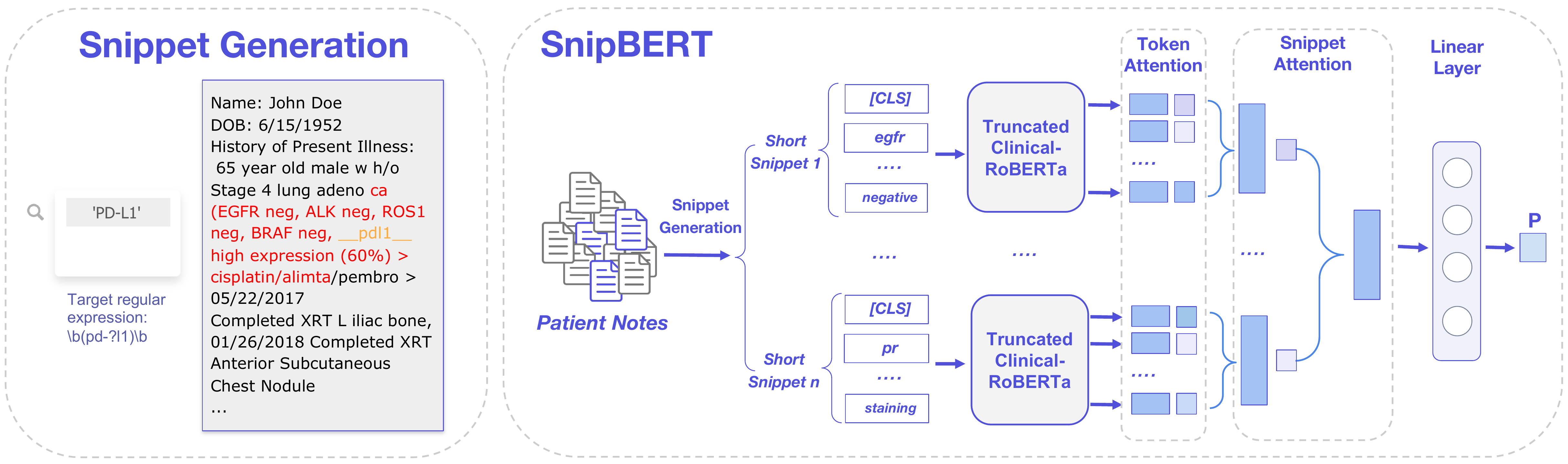}
    \vspace{-4mm}
    \caption{\mname method illustration.}
    \label{fig:method}
    \vspace{-3mm}
\end{figure*}

Recently, BERT-based methods have shown dominant results in many Natural Language Processing (NLP) tasks \citep{devlin2018bert}. Clinical versions of BERT have shown promising results in extracting information from short clinical text such as discharge summaries \citep{huang2019clinicalbert, mulyar2019phenotyping, valmianski2019evaluating}. However, there are challenges with using BERT for tasks that require processing long clinical text, such as an entire patient chart containing many clinical notes and reports. Notably, BERT has $O(n^2)$ memory/computational complexity, where $n$ is the number of input tokens, with a default maximum length of 512 tokens. While recent works have improved efficiency and applied BERT to text a few thousand words in length \citep{beltagy2020longformer}, EHRs can contain orders of magnitude more text than this. For example, in the real-world oncology database used in this study, patients had on average 265,000 words in their EHR.  Besides being computationally infeasible to process, much of this text may be irrelevant to a variable of interest, causing NLP models to produce poor and difficult-to-interpret results. 

\xhdr{Present work} We propose a new end-to-end finetuning approach called \mname to obtain efficient, interpretable predictions from long clinical text. Empirically, \mname achieves significantly better performance than TF-IDF and previous BERT-based approaches. We also show a case study of model interpretability.

\vspace{-6mm}
\section{Problem setting and dataset}
\vspace{-2mm}

A common pattern for information extraction problems in clinical text is to predict a categorical variable given a sequence of clinical notes associated with the patient. Here, we focus on three example tasks that can be important for downstream applications. The variables chosen are oncology-specific, though our methods are domain-general. The variables are: (1) \textbf{PD-L1 Biomarker (PD-L1)}: classify if a patient's tumor is tested for PD-L1 expression and if so, if the test result is positive, negative, or unknown/indeterminate; (2) \textbf{Next-Generation Sequencing (NGS)}: classify if a patient has undergone next-generation DNA sequencing; (3) \textbf{Metastatic Breast Cancer (mBC)}: classify if a patient has been diagnosed with metastatic breast cancer. We conducted our study on clinical notes sampled from the Flatiron Health database, a longitudinal, nationwide, de-identified database, derived from EHR data. Institutional Review Board approval of the study protocol was obtained prior to study conduct, and informed consent was waived. The PD-L1, NGS, mBC labels were abstracted by experts via chart reviews. The dataset includes 59,376 patients for PD-L1, 46,877 patients for NGS, and 46,724 patients for mBC.

\vspace{-5mm}
\section{Method} \label{sec:method}
\vspace{-2mm}

\mname consists of three stages: (1) Pretraining Clinical-RoBERTa; (2) Snippet Generation; (3) A Hierarchical Attention Snippet-Level Finetuning Approach.

\xhdr{Pretraining Clinical-RoBERTa}
To capture the characteristics of clinical text, we pretrained using text from 50,000 patients in the database (non-overlapping with the abstracted patient cohorts), which consists of 3,170,573 documents and 58 billion tokens. We then obtain a clinical text pretrained RoBERTa model ($\mathcal{F}$) \citep{liu2019roberta}.  

\xhdr{Snippet Generation} To eliminate irrelevant text and make long clinical text feasible to process, we use a snippet extraction technique to identify potentially-informative text for a given task. Using a set of task-specific search terms and/or regular expressions defined in collaboration with clinicians, we search through the text and, for each hit, extract a fixed-width snippet around the hit. Figure~\ref{fig:method} shows an example of using a PDL1 regular expression to extract a snippet (in red) from surrounding text.

\xhdr{A Hierarchical Attention Snippet-Level Finetuning Approach} Given a sequence of short snippets $\{S_1, \cdots, S_m\}$, we feed each snippet individually through $\mathcal{F}$. Note that $\mathcal{F}$ is truncated to set the maximum positions length as the snippet length $L$, which is much smaller than the default 512. For each snippet $S_i$, we obtain hidden representation $\{\mathbf{h}_i^1, \mathbf{h}_i^2, \cdots, \mathbf{h}_i^L\}$ where $\mathbf{h}_i^j$ is the embedding of $i$-th snippet's $j$-th token. To provide interpretability and improve performance, we adopt attention mechanisms on both the token and snippet level. In particular, we first apply a token-level attention layer which generates a score $C^j_{\mathrm{token}}$ for each token $j$ measuring its relevance. We then obtain a representation $\mathbf{h}_i$ for the snippet $i$ by taking the linear combination of tokens via $\mathbf{h}_i = \sum_j^L C^j_{\mathrm{token}} * \mathbf{h}_i^{j}$. We use a similar attention mechanism over snippets to produce a patient representation $\mathbf{h}$ via $\mathbf{h} = \sum_i^m C^i_{\mathrm{snip}} * \mathbf{h}_i$. The patient representation $\mathbf{h}$ is then fed into a linear decoder layer to generate a probability score $p_i$ for each class $i$. During training, we use cross-entropy loss to train the model end-to-end.

Note that \mname is very different from the standard BERT concatenation approach where individual sequences (snippets) are concatenated into a single document and fed into the model~\citep{huang2019clinicalbert,devlin2018bert,beltagy2020longformer,adhikari2019docbert}. In contrast, \mname processes each short snippet individually and aggregates them in an end-to-end manner. We observe two  beneficial properties over the concatenation approach.

\xhdr{Observation 1 (Complexity Reduction)}
For text with $m$ snippets of length $L$, the complexity of \mname is $m$ times smaller than the concatenation approach.

\xhdr{Explanation of Observation 1} The complexity of a transformer is $O(n^2)$. Then for the concatenation approach, the complexity is $\gamma (m*L)^2$. For the snippet-level approach, the complexity is $\gamma (L)^2 * m$. Then, $\frac{\gamma (L)^2 * m}{\gamma (m*L)^2} = m$. The complexity reduction makes it feasible for \mname to process much more text than the concatenation approach.

\xhdr{Observation 2 (Cross-snippet Interference)}
The snippet approach eliminates cross-snippet interference that could hurt performance.

\xhdr{Explanation of Observation 2} With a concatenation approach, distant snippets may be concatenated together. For example, a snippet from a patient's PD-L1 test results may be next to a snippet in which a physician considered ordering the test. This kind of juxtaposition is not present during pretraining, and the RoBERTa model may attend across snippets in unwanted ways, which we call ``cross-snippet interference.'' The snippet-level approach eliminates this possibility, and results in superior performance as confirmed in the Results section.
 
\begin{table*}[t]
    \centering
    \caption{\small Finetuning Tasks Predictive Performance.}
    \adjustbox{max width=0.76\textwidth}{%
    \begin{tabular}{l|l|ccc|ccc|ccc}
    \toprule
    \multicolumn{2}{c|}{Dataset} & \multicolumn{3}{c|}{\textbf{mBC}} & \multicolumn{3}{c|}{\textbf{NGS}} & \multicolumn{3}{c}{\textbf{PD-L1} (F1 score)}\\
    \toprule
    Setup & Model & {\small PR95} & {\small PR-AUC} & {\small ROC-AUC} & {\small PR95} & {\small PR-AUC} & {\small ROC-AUC}  & NEG & POS & {\small UNK}\\
    \midrule
    \multirow{2}{*}{TF-IDF} & TF-IDF-Full & 0.468 & 0.829 & 0.968 & 0.433 & 0.864 & 0.953 & 0.307 & 0.147 & 0.494 \\
    & TF-IDF-Snip & 0.742 & 0.925 & 0.987  & 0.729 & 0.945 & 0.983 & 0.761 & 0.658 & 0.708 \\
    \midrule
     \multirow{2}{*}{Concat} & RoBERTa & 0.429 & 0.871 & 0.969 & 0.389 & 0.818 & 0.935 & 0.729 & 0.627 & 0.695\\
    & Clinical-RoBERTa & 0.464 & 0.886 & 0.973&0.386 &0.823 & 0.938&0.695 &0.633 & 0.651 \\ 
    \midrule
    \multirow{2}{*}{Snippet} & RoBERTa & 0.824 & 0.955 & 0.991 & 0.835 & 0.952 & 0.986 & 0.784 & \textbf{0.698} & 0.728 \\
    & \mname (Ours) & \textbf{0.894} & \textbf{0.969} & \textbf{0.994} & \textbf{0.866} & \textbf{0.957} & \textbf{0.987} & \textbf{0.789} & 0.696 & \textbf{0.746} \\ 
    \bottomrule
    \end{tabular}
    }
    \label{tab:my_label}
    \vspace{-2mm}
\end{table*}

\begin{figure*}[t]
    \centering
    \includegraphics[width = 0.76\textwidth]{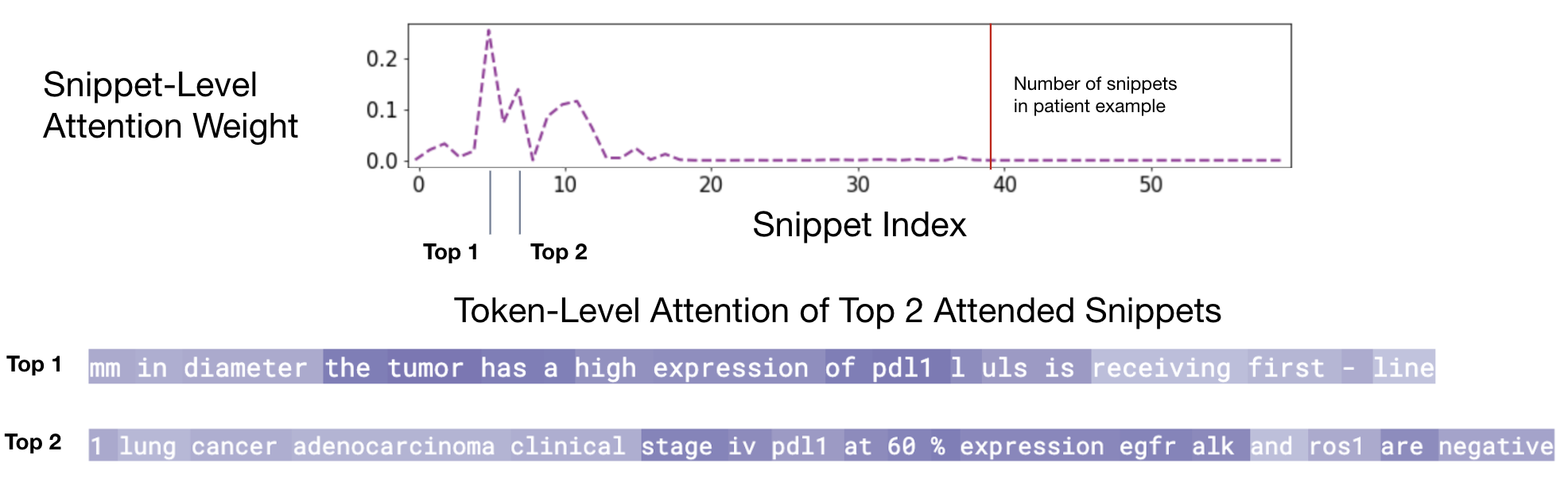}
    \vspace{-4mm}
    \caption{ \mname identifies informative snippets and tokens using hierarchical attention.}
    \label{fig:my_label}
    \vspace{-3mm}
\end{figure*}

\vspace{-3mm}
\section{Experiments}
\vspace{-2mm}

We conducted experiments to test the predictive power of the snippet-level approach and to explore the interpretability of the hierarchical attention mechanism.

We evaluate on the following methods. We use TF-IDF, which in some cases can outperform transformer models, as a strong baseline~\citep{valmianski2019evaluating}. We include two variants: TF-IDF-Full is applied on the full clinical notes for each patient and TF-IDF-Snip is applied on the generated snippets. Then, we include two methods, RoBERTa (without clinical domain pretraining) and Clinical-RoBERTa, applying both concatenation and snippet level approaches to each.  
 
Our evaluation metrics for binary variables mBC and NGS include PR-AUC, ROC-AUC, and Precision at Recall of 95\% (PR95). PR95 is useful for retrospective research or trials matching applications where the goal is to assemble a cohort that includes (nearly) all eligible patients for downstream processing. For PD-L1, we report F1 score for each of the negative (NEG), positive (POS) and unknown/indeterminate (UNK) class.

\vspace{-3mm}
\subsection{Results}
\vspace{-2mm}

Predictive results are shown in Table~\ref{tab:my_label}.

\xhdr{Snippet generation reduces noise} TF-IDF-Snip has significantly better predictive performance than TF-IDF-Full (up to 27.4\%), suggesting that snippet generation reduces noise and improves performance.

\xhdr{Snippet approach significantly outperforms concatenation approach} We observe that the snippet level approach improves up to a 43\% absolute increase in PR95 than concatenation counterparts and has a 15\% increase over the TFIDF baseline. These results support our observations that the snippet-level approach can improve not just efficiency but performance as well.

\xhdr{\mname provides clues for model interpretation} The snippet-level approach offers a simple way to see which snippets and tokens in the patient text led to a prediction. Figure~\ref{fig:my_label} shows an example for a PD-L1 positive patient. The snippet level attentions are sparse as the model mainly weights just 5 out of 39 snippets to make the prediction. Within the top 2 snippets, the token attentions directly flag the high expression (i.e., positivity) of PD-L1. This example shows the potential explainability of \mname. Note that interpretability is infeasible for concatenation and general BERT-based approaches since they lack the ability to generate attention on the snippet level.

\xhdr{Pretraining improves performance} For pretraining, the masked language modeling loss is 2.664 (6.97\% accuracy) for RoBERTa-base and 1.175 (30.88\% accuracy) for Clinical-RoBERTa. We observe the pretrained model usually outperforms the base model in all three downstream tasks. 

\vspace{-3mm}
\section{Conclusion}
\vspace{-2mm}

We propose a novel BERT-based finetuning approach to unlock valuable information in long clinical texts. It significantly improves predictive performance and efficiency while also providing explainability. For future work, we plan to quantitatively examine interpretability and explore the performance of \mname across a broader array of tasks.

\bibliography{jmlr-sample}

\begin{thebibliography}{12}
\providecommand{\natexlab}[1]{#1}
\providecommand{\url}[1]{\texttt{#1}}
\expandafter\ifx\csname urlstyle\endcsname\relax
  \providecommand{\doi}[1]{doi: #1}\else
  \providecommand{\doi}{doi: \begingroup \urlstyle{rm}\Url}\fi

\bibitem[Adhikari et~al.()Adhikari, Ram, Tang, Hamilton, and
  Lin]{adhikari2019docbert}
Ashutosh Adhikari, Achyudh Ram, Raphael Tang, William~L. Hamilton, and Jimmy
  Lin.
\newblock Exploring the limits of simple learners in knowledge distillation for
  document classification with {D}oc{BERT}.
\newblock In \emph{Proceedings of the 5th Workshop on Representation Learning
  for NLP}, pages 72--77.

\bibitem[Beltagy et~al.(2020)Beltagy, Peters, and Cohan]{beltagy2020longformer}
Iz~Beltagy, Matthew~E Peters, and Arman Cohan.
\newblock Longformer: The long-document transformer.
\newblock \emph{arXiv preprint arXiv:2004.05150}, 2020.

\bibitem[Boag et~al.(2018)Boag, Doss, Naumann, and Szolovits]{boag2018s}
Willie Boag, Dustin Doss, Tristan Naumann, and Peter Szolovits.
\newblock What’s in a note? unpacking predictive value in clinical note
  representations.
\newblock \emph{AMIA Summits on Translational Science Proceedings},
  2018:\penalty0 26, 2018.

\bibitem[Devlin et~al.(2018)Devlin, Chang, Lee, and Toutanova]{devlin2018bert}
Jacob Devlin, Ming-Wei Chang, Kenton Lee, and Kristina Toutanova.
\newblock Bert: Pre-training of deep bidirectional transformers for language
  understanding.
\newblock \emph{NAACL}, 2018.

\bibitem[Huang et~al.(2019)Huang, Altosaar, and
  Ranganath]{huang2019clinicalbert}
Kexin Huang, Jaan Altosaar, and Rajesh Ranganath.
\newblock Clinicalbert: Modeling clinical notes and predicting hospital
  readmission.
\newblock \emph{CHIL Workshop}, 2019.

\bibitem[Khozin et~al.(2017)Khozin, Blumenthal, and Pazdur]{khozin2017real}
Sean Khozin, Gideon~M Blumenthal, and Richard Pazdur.
\newblock Real-world data for clinical evidence generation in oncology.
\newblock \emph{JNCI: Journal of the National Cancer Institute}, 109\penalty0
  (11):\penalty0 djx187, 2017.

\bibitem[Liu et~al.(2018)Liu, Zhang, and Razavian]{liu2018deep}
Jingshu Liu, Zachariah Zhang, and Narges Razavian.
\newblock Deep ehr: Chronic disease prediction using medical notes.
\newblock \emph{MLHC}, 2018.

\bibitem[Liu et~al.(2019)Liu, Ott, Goyal, Du, Joshi, Chen, Levy, Lewis,
  Zettlemoyer, and Stoyanov]{liu2019roberta}
Yinhan Liu, Myle Ott, Naman Goyal, Jingfei Du, Mandar Joshi, Danqi Chen, Omer
  Levy, Mike Lewis, Luke Zettlemoyer, and Veselin Stoyanov.
\newblock Roberta: A robustly optimized bert pretraining approach.
\newblock \emph{arXiv preprint arXiv:1907.11692}, 2019.

\bibitem[Mulyar et~al.(2019)Mulyar, Schumacher, Rouhizadeh, and
  Dredze]{mulyar2019phenotyping}
Andriy Mulyar, Elliot Schumacher, Masoud Rouhizadeh, and Mark Dredze.
\newblock Phenotyping of clinical notes with improved document classification
  models using contextualized neural language models.
\newblock \emph{NeurIPS ML4H}, 2019.

\bibitem[Shickel et~al.(2017)Shickel, Tighe, Bihorac, and
  Rashidi]{shickel2017deep}
Benjamin Shickel, Patrick~James Tighe, Azra Bihorac, and Parisa Rashidi.
\newblock Deep ehr: a survey of recent advances in deep learning techniques for
  electronic health record (ehr) analysis.
\newblock \emph{IEEE journal of biomedical and health informatics}, 22\penalty0
  (5):\penalty0 1589--1604, 2017.

\bibitem[Valmianski et~al.(2019)Valmianski, Goodwin, Finn, Khan, and
  Zisook]{valmianski2019evaluating}
Ilya Valmianski, Caleb Goodwin, Ian~M Finn, Naqi Khan, and Daniel~S Zisook.
\newblock Evaluating robustness of language models for chief complaint
  extraction from patient-generated text.
\newblock \emph{NeurIPS ML4H}, 2019.

\bibitem[Weng et~al.(2017)Weng, Wagholikar, McCray, Szolovits, and
  Chueh]{weng2017medical}
Wei-Hung Weng, Kavishwar~B Wagholikar, Alexa~T McCray, Peter Szolovits, and
  Henry~C Chueh.
\newblock Medical subdomain classification of clinical notes using a machine
  learning-based natural language processing approach.
\newblock \emph{BMC medical informatics and decision making}, 17\penalty0
  (1):\penalty0 1--13, 2017.

\end{thebibliography}

\appendix

\end{document}